\documentclass[conference]{IEEEtran}
\IEEEoverridecommandlockouts

\usepackage{cite}

\usepackage{amsmath,amssymb,amsfonts}
\usepackage{algorithmic}
\usepackage{graphicx,booktabs,multirow}
\usepackage{textcomp}
\usepackage{xcolor}
\def\BibTeX{{\rm B\kern-.05em{\sc i\kern-.025em b}\kern-.08em
    T\kern-.1667em\lower.7ex\hbox{E}\kern-.125emX}}
\begin{document}

\makeatletter
\newcommand{\linebreakand}{%
  \end{@IEEEauthorhalign}
  \hfill\mbox{}\par
  \mbox{}\hfill\begin{@IEEEauthorhalign}
}
\makeatother

\title{P$^3$-SAM: SAM with Perceptual Parallel Prompt for Few-Shot Strip Steel Surface Defect Segmentation
\thanks{$^*$Corresponding author. 
This work was supported in part by the the National Natural Science Foundation of China under Grant 62471278, and in part by the Research Grants Council of the Hong Kong Special Administrative Region, China under Grant STG5/E-103/24-R.}
}

\author{\IEEEauthorblockN{Qian Xu}
\IEEEauthorblockA{
\textit{Shandong University}\\
Jinan, China \\
lsle.v.epl@gmail.com}
\and

\IEEEauthorblockN{Hang Xiong}
\IEEEauthorblockA{
\textit{Alibaba International Digital Commercial Group} \\
Beijing, China \\
xionghang.xh@alibaba-inc.com}
\and

\IEEEauthorblockN{Anpeng Wang$^{*}$}
\IEEEauthorblockA{
\textit{Shandong University}\\
Jinan, China \\
rawwap@mail.sdu.edu.cn}
\linebreakand

\IEEEauthorblockN{Sam Kwong}
\IEEEauthorblockA{
\textit{Lingnan University}\\
Hong Kong, China \\
samkwong@ln.edu.hk}
\and

\IEEEauthorblockN{Cong Zhang}
\IEEEauthorblockA{
\textit{Shandong University}\\
Jinan, China \\
congzhang@sdu.edu.cn}
\and

\IEEEauthorblockN{Runmin Cong$^{*}$}
\IEEEauthorblockA{
\textit{Shandong University}\\
Jinan, China \\
rmcong@sdu.edu.cn}

}

\maketitle
\begin{abstract}
Few-shot semantic segmentation (FSS) of strip steel surface defects (S$^3$D) has posed significant challenges distinct from natural scenes. Unlike natural images, S$^3$D task exhibits unique characteristics including low local contrast, uneven illumination, and complex fine-grained texture patterns. 
Although recent methods based on Segment Anything Model (SAM) have shown promise in FSS on natural images by leveraging SAM's powerful pre-trained representations, these unique industrial characteristics of S$^3$D images lead to performance drop when directly applying SAM to industrial defect scenarios. 
In this paper, we propose a novel Perceptual Parallel Prompt (P$^3$) framework that empowers SAM, creating the P$^3$-SAM model to address these challenges through two core strategies.
First, we develop a Perceptual-Optimized Encoding (POE) strategy that enhances local contrast and preserves critical texture details for S$^3$D segmentation. 
Second, we introduce the Parallel Prompt Generator (PPG) strategy that simultaneously generates both semantic and spatial prompts, enabling comprehensive guidance for SAM's decoder across varying images. 
Extensive experiments on three few-shot S$^3$D benchmarks demonstrate that P$^3$-SAM achieves state-of-the-art performance, with particularly notable improvements of 12.00\% in mIoU on Surface Defects-4i dataset.
\end{abstract}

\begin{IEEEkeywords}
few-shot semantic segmentation, strip steel surface defects segmentation, segment anything model, prompt generation
\end{IEEEkeywords}

\begin{figure}[!t]
\centering
\includegraphics[width=1.0\linewidth]{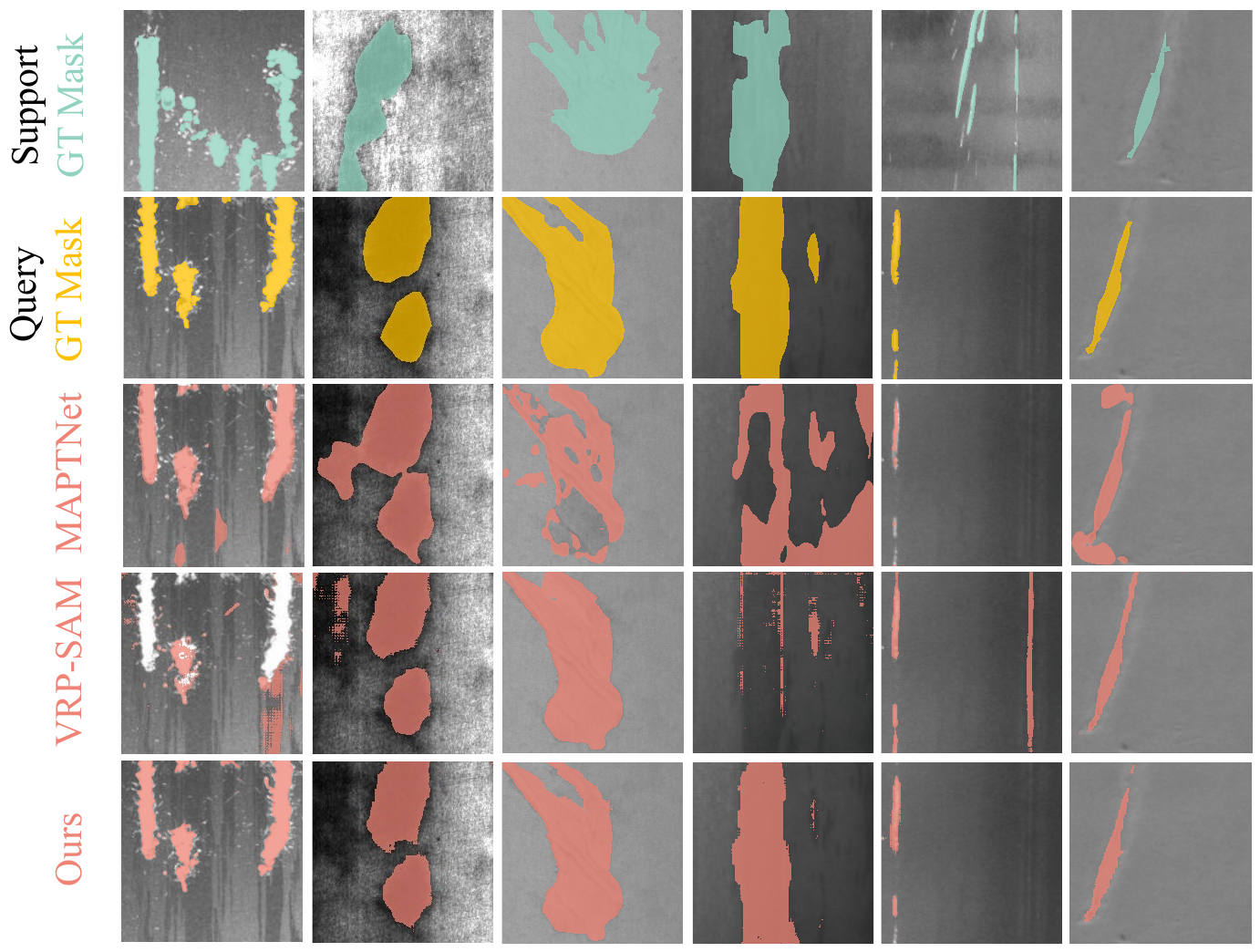}
\caption{\textbf{Visualization of comparative results of Ours with MAPTNet and VRP-SAM.} From top to bottom, each row illustrates the ground truth masks of support images in green, the ground truth masks of query images in yellow, and the predictions generated by various methods in red.}
\label{fig:motivation}
\end{figure}

\section{Introduction}
Semantic segmentation~\cite{r01,r02}, which assigns pixel-level class labels to images, has become a fundamental task in computer vision. While deep learning-based approaches have achieved remarkable success, they heavily rely on large-scale annotated datasets that are costly and time-consuming to collect, especially in specialized domains. 
Few-shot semantic segmentation (FSS)~\cite{r03} addresses this data scarcity challenge by leveraging meta-learning principles~\cite{r08}, aiming to segment novel classes with only a few annotated support samples via an episodic training strategy. 
Despite progress on natural image benchmarks~\cite{r09,r10}, FSS still struggles with feature distribution shift between training and testing classes, limited semantic information from few supports, and difficulty matching support-query features under appearance variations.

These challenges become even more pronounced when applying FSS to strip steel surface defect (S$^3$D) segmentation \cite{r12}, where industrial defect images differ significantly from natural scenes.
As shown in Fig.~\ref{fig:motivation}, S$^3$D images exhibit monotonous color distributions but suffer from low local contrast and uneven illumination. Moreover, defect regions are often obscured by complex textures with subtle intensity variations, hindering precise localization of small defects.
Additionally, unlike natural scenes that rely on high-level semantic features, the segmentation of S$^3$D depends on fine-grained structural and textural details. The critical cues lie in local intensity variations and micro-textures.
Existing FSS methods \cite{r14}, however, typically extract deep-layer features and discard these critical low-level details. 
This raises our first research question: \textit{How can we encode feature representations to address the unique challenges posed by S$^3$D characteristics under few-shot constraints?}

Existing FSS methods predominantly adopt prototype-based architectures \cite{r14}, which extract prototypes and perform pixel-wise matching for segmentation.
As illustrated in Fig.~\ref{fig:motivation}, these approaches (e.g., MAPTNet \cite{r14}) fail to fully exploit features for S$^3$D segmentation, particularly struggling with low contrast and texture preservation. 
Recently, the Segment Anything Model (SAM) \cite{r16}, a powerful foundation model trained on billions of masks, accepts diverse prompts to guide segmentation. Recent SAM-based FSS methods \cite{r18} have achieved impressive performance on natural image benchmarks. However, applying SAM to few-shot S$^3$D scenarios faces critical prompting challenges. SAM's original design requires user-provided prompts, which is impractical in few-shot settings. 
To address this, existing SAM-based methods \cite{r18} generate augmented features and sample sparse prompts to guide SAM. 
However, this approach discards rich spatial information and exhibits high sensitivity to feature quality, making it unsuitable for S$^3$D segmentation under few-shot constraints. This raises our second research question: \textit{How can we provide effective prompts to guide SAM for few-shot S$^3$D segmentation?}

To address these questions, we propose a novel Perceptual Parallel Prompt (P$^3$) framework that empowers SAM, creating the P$^3$-SAM model to tackle the unique challenges of few-shot S$^3$D segmentation through two core strategies.
First, we develop a Perceptual-Optimized Encoding (POE) strategy that enhances local contrast and preserves critical texture details. Specifically, POE unifies defect representation through perceptual optimization via multi-scale Retinex and augments shallow features with prototype and pseudo-mask guidance to capture fine-grained textural characteristics.
Second, we introduce a Parallel Prompt Generator (PPG) strategy that simultaneously generates both semantic and spatial prompt embeddings, enabling comprehensive guidance for SAM's decoder across varying images. Unlike existing methods relying solely on sparse prompts, PPG generates both prompt types in parallel from enhanced features, allowing complete spatial structure inference. As illustrated in Fig.~\ref{fig:framework}, our framework effectively addresses the unique challenges of few-shot S$^3$D segmentation. In summary, our contributions are:

\begin{itemize}
  \item We propose a novel P$^3$-SAM model that achieves state-of-the-art performance on three few-shot S$^3$D benchmarks, notably improving 12.00\% mIoU on Surface Defects-4i.
  \item We develop a POE strategy that combines perceptual optimization with prototype and mask guided learning to encode features suited for few-shot S$^3$D segmentation.
  \item We introduce a PPG strategy that generates semantic and spatial prompts in parallel, maintaining effective guidance across varying images.
\end{itemize}

\begin{figure*}[t]
    \centering
    \includegraphics[width=\linewidth]{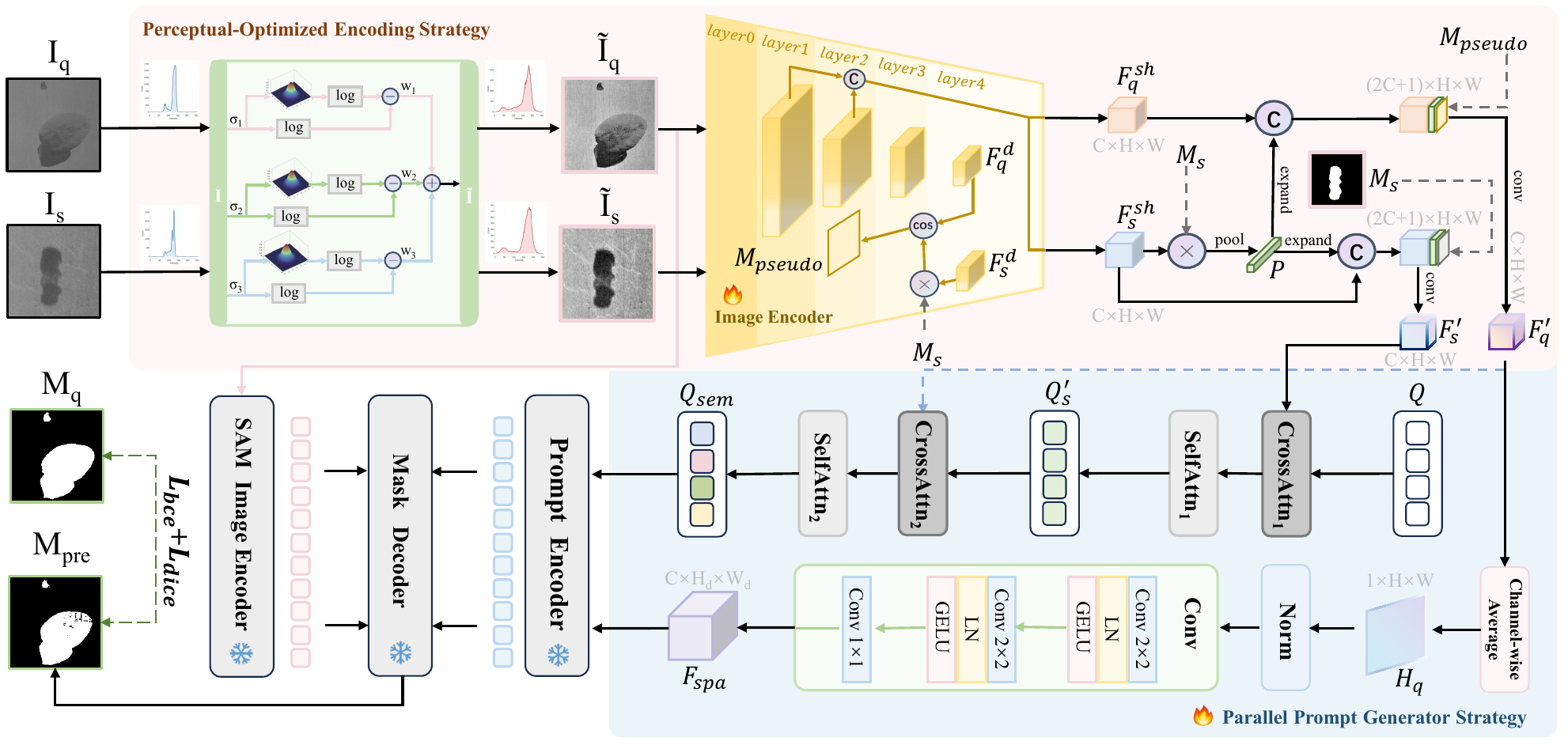}
    \caption{\textbf{Overview of the proposed P$^3$-SAM framework.} Our method consists of two core strategies: (1) Perceptual-Optimized Encoding (POE) strategy that unifies defect representation through perceptual optimization and augments shallow features with prototype and mask guided learning, and (2) Parallel Prompt Generator (PPG) strategy that generates both semantic and spatial prompts in parallel to provide comprehensive guidance for SAM's decoder.}
    \label{fig:framework}
\end{figure*}

\section{RELATED WORK}
\subsection{Few-Shot Semantic Segmentation}
Few-shot semantic segmentation (FSS), pioneered by Shaban et al. \cite{r19}, targets dense masks for novel classes from few labeled supports.
Prototype-based methods \cite{r14,ref5,r37} segment by matching queries to support prototypes incorporating various improvements \cite{r20,ref1,ref2}. 
Affinity-based methods \cite{r21,r22,ref4} construct dense correspondences between query and support features through feature concatenation with learnable CNN \cite{r01} or Transformer \cite{r23} modules for mask prediction. 
Foundation-based methods such as SAM \cite{r16} use well learned knowledge to simplify the learning of FSS and partially mitigate class shift through promptable segmentation.
Despite progress, shallow structural cues crucial for S$^3$D remain underused. To address low contrast, uneven illumination, strong textures, and slender defects, we retain these cues in our work under few-shot constraints to improve robustness.

\subsection{Application of SAM}
Existing SAM-based methods fall into two categories. The first leverages SAM outputs as priors for downstream tasks, such as weakly supervised pseudo-labeling and data annotation(SEPL). This strategy is modular but depends on zero-shot generalization and can amplify prior errors in challenging industrial scenes.
The second guides SAM's segmentation through various prompt combinations\cite{r25,ref3}. Recent work VRP-SAM \cite{r18} extends SAM to visual reference \cite{r25,r26} prompting by using annotated support images. However, it struggles when reference annotations are sparse or incomplete, particularly under low contrast and strong background textures in few-shot S$^3$D settings, as sparse prompts provide weak spatial constraints while dense prompts degenerate without complete masks. 
We address these issues by generating both semantic and spatial prompts in parallel, which strengthens comprehensive guidance for SAM across varying images under the constraint of limited support annotations.

\section{METHOD}
\subsection{Problem Definition}
FSS aims to segment target objects in a query image given only a few annotated support examples. Following the meta-learning paradigm, the model is trained using episodes rather than conventional image batches. Each episode consists of a support set and a query set. In a $K$-shot setting, the support set $S = \{(I_s^i, M_s^i)\}_{i=1}^{K}$ includes $K$ support images $I_s$ and their corresponding binary masks $M_s$, while the query set $Q = \{(I_q, M_q)\}$ comprises a query image $I_q$ and its ground-truth mask $M_q$ used for loss calculation during training. 

The episodes are sampled from a training dataset $\mathcal{D}_{\text{train}} = \{(S^i, Q^i)\}_{i=1}^{N_{\text{train}}}$ for meta-training and from a testing dataset $\mathcal{D}_{\text{test}} = \{(S^i, Q^i)\}_{i=1}^{N_{\text{test}}}$ for meta-testing. $\mathcal{D}_{\text{train}}$ contains object classes $\mathcal{C}_{\text{train}}$, and $\mathcal{D}_{\text{test}}$ contains object classes $\mathcal{C}_{\text{test}}$, with no overlap between them, i.e., $\mathcal{C}_{\text{train}} \cap \mathcal{C}_{\text{test}} = \emptyset$. The objective is to train the model on $\mathcal{D}_{\text{train}}$ and evaluate its generalization on unseen classes in $\mathcal{D}_{\text{test}}$, following the learn-to-learn paradigm. 

\subsection{Overview}
Our proposed P$^3$-SAM addresses the challenges of few-shot S$^3$D segmentation through two core strategies, as illustrated in Fig.~\ref{fig:framework}. 
First, we develop a Perceptual-Optimized Encoding (POE) strategy that unifies defect representation through perceptual optimization via pixel-level multi-scale Retinex, then extracts fine-grained texture information from shallow backbone layers and augments them with prototype and mask guided learning to preserve critical low-level details. 
Second, we introduce the Parallel Prompt Generator (PPG) strategy that generates both semantic and spatial prompt embeddings in parallel from the enhanced query and support features. This enables comprehensive guidance for SAM's decoder. The enhanced features from POE are processed by PPG to produce parallel prompts, which are then fed into SAM's frozen prompt encoder and decoder for final segmentation.

\subsection{Perceptual-Optimized Encoding Strategy}
Unlike natural images where high-level semantic features dominate segmentation, S$^3$D tasks rely heavily on low-level texture and edge information due to subtle defect patterns and fine-grained surface variations. Moreover, uneven illumination and low local contrast further obscure fine-grained defect textures, making precise localization difficult. To address these challenges, we propose the POE strategy to enhance feature representations for few-shot S$^3$D segmentation.

We first employ pixel-level multi-scale Retinex \cite{r27} for perceptual optimization, which decomposes the input image into reflectance and illumination components, suppressing uneven illumination while enhancing local contrast and texture details. In the logarithmic domain, for each RGB channel, the single-scale response at scale $\sigma$ is computed as:
\begin{equation}
R_{\sigma}(x) = \log I(x) - \log\bigl( I(x) \ast G_{\sigma}(x) \bigr),
\end{equation}
where $G_{\sigma}$ is a Gaussian kernel and $\ast$ denotes convolution. The multi-scale responses at three scales $\sigma \in \{15, 80, 250\}$ are combined to capture different spatial frequencies, then mapped back to the linear domain and normalized to $[0, 255]$. The enhanced image $\tilde{I}$ is then fed into both the image encoder and SAM image encoder. This preprocessing is applied identically to both support images $\{I_s^i\}_{i=1}^{K}$ and query image $I_q$ during training and inference.

With the image encoder, we separately encode the enhanced support images $\{\tilde{I}_s^i\}_{i=1}^K$ and query image $\tilde{I}_q$ to extract both shallow and deep features. For shallow features $F^{\mathrm{sh}}$, we obtain $F_1 \in \mathbb{R}^{C_1 \times H_1 \times W_1}$ from layer1 (stride 4) and $F_2 \in \mathbb{R}^{C \times H \times W}$ from layer2 (stride 8). These multi-scale shallow features are combined:
\begin{equation}
F^{\mathrm{sh}} = \mathrm{Conv}\bigl(\mathrm{concat}\bigl(\mathrm{Up}(F_1), F_2\bigr)\bigr) \in \mathbb{R}^{C \times H \times W},
\end{equation}
where $C$ is the target channel dimension and $H \times W$ matches the layer2 resolution. To inject semantic guidance, we compute a class-specific prototype from the shallow features via masked average pooling:
\begin{equation}
P^i = \mathrm{MaskAvgPool}\bigl(F^{\mathrm{sh}}_s, M_s^i\bigr)
\end{equation}
For the query image, we extract deep features $F^{d}$ from layer3-4 and generate a pseudo-mask $M_{\mathrm{pseudo}}$:
\begin{equation}
M_{\mathrm{pseudo}} = \mathrm{CosSim}\bigl(F_s^d \otimes M_s, F_q^d\bigr),
\end{equation}
where $\otimes$ denotes element-wise multiplication and $\mathrm{CosSim}(\cdot)$ computes the cosine similarity. Finally, the shallow features are concatenated with the prototype and corresponding masks, followed by convolution for feature enhancement:
\begin{align}
F_s' &= \mathrm{Conv}\bigl( \mathrm{concat}( F_s^{\mathrm{sh}}, P^i, M_s^i ) \bigr), \\
F_q' &= \mathrm{Conv}\bigl( \mathrm{concat}( F_q^{\mathrm{sh}}, P^i, M_{\mathrm{pseudo}} ) \bigr),
\end{align}
where $\mathrm{concat}(\cdot)$ denotes concatenation. By combining multi-scale Retinex optimization with prototype and mask guided shallow feature augmentation, the POE strategy enhances local contrast and texture details while preserving fine-grained information from shallow layers. This design achieves a balance between low-level texture preservation and high-level semantic guidance, enabling the enhanced features $F_s', F_q' \in \mathbb{R}^{C \times H \times W}$ to capture both critical texture details and class-specific context for subsequent prompt generation.

\begin{table*}[htbp]
\caption{Performance Comparison of mIoU and FB-IoU on FSSD-12, Surface Defects-4i and ESDIs-SOD datasets.}
\begin{center}
\label{tab:comparison}
\begin{tabular}{ccccccccccccc}
    \toprule
    \multirow{2.5}*{Dataset} & \multirow{2.5}*{Method} &  \multirow{2.5}*{Backbone}  & \multicolumn{4}{c}{mIoU(1-shot)} & \multirow{2.5}*{FB-IoU} & \multicolumn{4}{c}{mIoU(5-shot)} & \multirow{2.5}*{FB-IoU}\\
     \cmidrule(lr){4-7} \cmidrule(lr){9-12} &&& Fold-0 & Fold-1 & Fold-2 & MEAN &  & Fold-0 & Fold-1 & Fold-2 & MEAN \\
    \midrule
    \multirow{7}*{FSSD-12} 
        & TGRNet    & \multirow{7}*{ResNet50} & 61.80 & 62.00 & 48.30 & 57.70 & 73.60 & 62.40 & 59.70 & 47.80 & 58.50 &75.10\\
        & CPANet    && 66.00 & 64.00 &54.60 & 61.50 &76.10 & 66.50 & 64.90 & 56.30 & 62.60 & 76.30\\
        & SCCAN    && 53.70 & 60.70 & 50.90 & 55.10 & 70.70 & 56.90 & 61.10 & 53.40 & 57.10 & 72.20\\
        & DCP    && 58.30 & 51.70 & 43.10 & 51.00 & 70.60 & 55.90 & 47.00 & 55.70 & 52.90 & 70.90\\
        & MAPTNet    && 69.70 & 71.30 & 58.60 & 66.50 & 80.00 & 69.00 & 72.30 & \underline{62.60} & 68.00 & 80.60\\
        & VRP-SAM      && \underline{70.39} & \textbf{82.64} & \underline{62.44} & \underline{71.82} & \underline{82.31} & \underline{70.72} & \textbf{82.48} & 61.97 & \underline{71.72} & \underline{82.27} \\
        & Ours     && \textbf{73.97} & \underline{81.61} & \textbf{69.77} & \textbf{75.12} & \textbf{85.09} & \textbf{73.62} & \underline{81.76} & \textbf{70.96} & \textbf{75.45} & \textbf{85.31}\\
        \cmidrule(lr){1-13}
    \multirow{7}*{Surface Defects-4i}
        & TGRNet    & \multirow{7}*{ResNet50}  & 45.10 & 30.10 & 23.20 & 32.80 & 55.90 & 44.10 & 35.00 & 23.80 & 34.80 & 53.90\\
        & CPANet    && 38.00 & 31.80 & 24.70 & 31.50 & 54.80 & 38.40 & 40.30 & 24.80 & 34.50 & 57.10\\
        & SCCAN     && 28.30 & 37.90 & 27.10 & 31.10 & 54.80 & 27.40 & 39.30 & 26.50 & 31.10 & 55.20\\
        & DCP      && 28.90 & 38.40 & 25.20 & 31.00 & 54.70 & 36.10 & 43.30 & 28.40 & 36.00 & 59.00\\
        & MAPTNet      && 41.20 & 43.20 & 25.20 & 36.40 & 57.40 & \underline{50.80} & 46.60 & 29.30 & 42.20 & 61.00\\
        & VRP-SAM      && \underline{45.57} & \underline{47.73} & \underline{36.92} & \underline{43.41} & \underline{62.79} & 45.47 & \underline{47.06} & \underline{36.70} & \underline{43.08} & \underline{62.40}\\
        & Ours     && \textbf{55.80} & \textbf{49.22} & \textbf{61.20} & \textbf{55.41} & \textbf{69.54} & \textbf{54.90} & \textbf{49.69} & \textbf{61.20} & \textbf{55.26} & \textbf{69.18}\\
        \cmidrule(lr){1-13}
    \multirow{7}*{ESDIs-SOD}
        & TGRNet       & \multirow{7}*{ResNet50}& 53.60 & \underline{70.00} & 42.40 & 55.30 & 72.50 & 53.60 & 72.70 & 47.70 & 58.00 & 75.00\\
        & CPANet        && 49.40 & 69.40 & 42.10 & 53.60 & 72.00 & 51.90 & 70.50 & 44.50 & 55.60 & 72.60\\
        & SCCAN     && 53.10 & \textbf{72.90} & 41.90 & 56.00 & \underline{72.90} & 50.60 & \underline{73.40} & 44.50 & 56.20 &72.90\\
        & DCP       && 53.70 & 69.10 & 42.80 & 55.20 & 72.30 & 54.20 & 71.10 & 45.90 & 57.00 & 74.70\\
        & MAPTNet       && 50.40 & 68.50 & 44.00 & 54.30 & 72.50 & 54.30 & \textbf{74.70} & 50.50 & 59.80 & \underline{76.20}\\
        & VRP-SAM      && \underline{60.72} & 60.11 & \underline{60.92} & \underline{60.58} & 71.25 & \underline{60.94} & 60.60 & \underline{60.95} & \underline{60.83} & 71.31\\
        & Ours     && \textbf{61.67} & 62.24 & \textbf{61.83} & \textbf{61.91} & \textbf{74.09} & \textbf{62.24} & 62.89 & \textbf{62.30} & \textbf{62.48} & \textbf{76.93}\\       
    \bottomrule
  \end{tabular}
\end{center}
\end{table*}

\subsection{Parallel Prompt Generator Strategy}
Existing SAM-based FSS methods typically generate semantic-only prompts which lack spatial structural information, causing the prompt embedding to collapse to a degenerate constant solution.
As illustrated in Fig.~\ref{fig:moti2}, while the SAM image encoder captures general visual features, relying solely on semantic prompts fails to provide explicit spatial structure guidance, resulting in incomplete segmentation. 
In contrast, incorporating both semantic and spatial prompts enables comprehensive guidance. 
To leverage both prompt types, we propose the Parallel Prompt Generator (PPG) strategy, which generates both semantic and spatial prompt embeddings in parallel from the enhanced features, enabling complete spatial structure inference with sustained effectiveness.

Given the enhanced support and query features $F_s', F_q' \in \mathbb{R}^{C \times H \times W}$ from POE, we first introduce a set of learnable queries $Q \in \mathbb{R}^{N \times C}$, where $N$ is the number of visual prompts. These queries interact with support features to obtain category-specific information through cross-attention and self-attention:
\begin{equation}
Q_s' = \mathrm{SelfAttn}_1\bigl( \mathrm{CrossAttn}_1(Q, F_s') \bigr),
\end{equation}
where $Q_s' \in \mathbb{R}^{N \times C}$ encodes knowledge of the defect category to be segmented. Subsequently, we employ cross-attention to interact these queries with query features to obtain foreground information:
\begin{equation}
Q_{\mathrm{sem}} = \mathrm{SelfAttn}_2\bigl( \mathrm{CrossAttn}_2(Q_s', F_q') \bigr),
\end{equation}
where $Q_{\mathrm{sem}} \in \mathbb{R}^{N \times C}$ serves as the semantic prompt embedding for the query, encoding class-specific semantic guidance from support images.

To provide spatial structure guidance, we generate a dense prompt embedding from query features. We first compress the multi-channel query features into a single-channel spatial heatmap via channel averaging:
\begin{equation}
H_q = \frac{1}{C} \sum_{c=1}^{C} F_q'(c) \in \mathbb{R}^{H \times W}
\end{equation}
This heatmap is then normalized to match SAM's prompt encoder input resolution, followed by convolutional encoding to generate the spatial prompt embedding:
\begin{equation}
F_{\mathrm{spa}} = \mathrm{Conv}\bigl( \mathrm{Norm}(H_q) \bigr) \in \mathbb{R}^{C \times H_d \times W_d},
\end{equation}
where $F_{\mathrm{spa}}$ provides spatial structure guidance derived from the query image itself, enabling it to adapt flexibly to variations in support annotation quality.

The final segmentation mask $M_{\mathrm{pre}} \in \mathbb{R}^{H_0 \times W_0}$ is produced by SAM's mask decoder, which integrates the image embedding $F_{\mathrm{base}}$ from SAM's frozen encoder, the semantic prompt embedding from SAM's prompt encoder processing $Q_{\mathrm{sem}}$ and the spatial prompt embedding from SAM's prompt encoder processing $F_{\mathrm{spa}}$:
\begin{equation}
M_{\mathrm{pre}} = \mathrm{Decoder}\bigl( F_{\mathrm{base}}, \mathrm{P}(Q_{\mathrm{sem}}), \mathrm{P}(F_{\mathrm{spa}}) \bigr),
\end{equation}
where $\mathrm{Decoder}(\cdot)$ denotes SAM's mask decoder and $\mathrm{P}(\cdot)$ denotes SAM's prompt encoder.
By generating semantic and spatial prompts in parallel, PPG combines semantic guidance via $Q_{\mathrm{sem}}$ and spatial structure via $F_{\mathrm{spa}}$ simultaneously. This parallel design enables the model to infer complete spatial structures from limited support annotations across varying images, achieving more accurate segmentation than relying on either prompt type alone.

\begin{figure}[t]
    \centering
    \includegraphics[width=\linewidth]{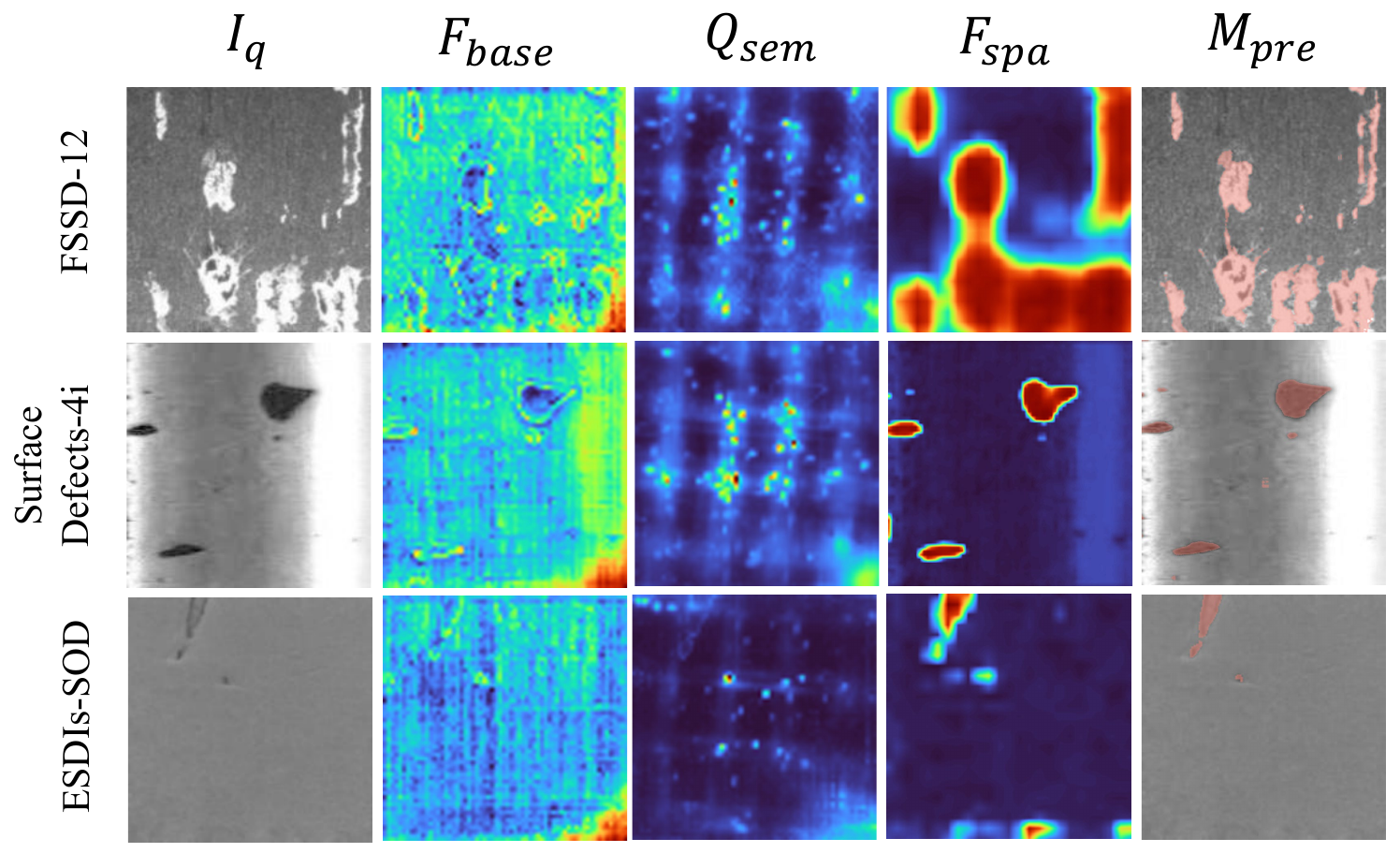}
    \caption{\textbf{Qualitative effect of parallel prompts on S$^3$D segmentation.} From left to right: query image $I_q$, feature $F_{\mathrm{base}}$ captured from SAM image encoder, semantic prompt embedding $Q_{\mathrm{sem}}$, spatial prompt embedding $F_{\mathrm{spa}}$, and final segmentation mask $M_{\mathrm{pre}}$.}
    \label{fig:moti2}
\end{figure}

\subsection{Loss Function}
We employ the Binary Cross-Entropy (BCE) loss and Dice loss \cite{r30,r31} to supervise the training of the entire pipeline. The loss between the final predicted mask $M_{\mathrm{pre}}$ and the ground-truth mask $M_{\mathrm{q}}$ for the query image is defined as:
\begin{equation}
L_{\mathrm{total}} = L_{\mathrm{bce}}(M_{\mathrm{pre}}, M_q) + L_{\mathrm{dice}}(M_{\mathrm{pre}}, M_q)
\end{equation}
where $L_{\mathrm{bce}}$ and $L_{\mathrm{dice}}$ are the BCE loss and Dice loss, respectively. The BCE loss ensures pixel-level accuracy, while the Dice loss provides additional spatial context by emphasizing region overlap. By combining both losses, we jointly consider accuracy and spatial coherence, enabling the model to produce more precise segmentation results.

\begin{table}[htbp]
    \caption{Ablation Study of Different Strategies on FSSD-12.}
    \label{tab:ablation}
    \centering
    \begin{tabular}{cccccc}
    \toprule
    \multirow{2.5}*{POE} & \multirow{2.5}*{PPG} & \multicolumn{2}{c}{1-shot} & \multicolumn{2}{c}{5-shot}\\
    \cmidrule(lr){3-6} && mIoU & FB-IoU & mIoU & FB-IoU\\
    \midrule
     & & 71.82  & 82.31 & 71.72 & 82.27\\    
     \checkmark & & 73.92 & 84.07 & 74.01 & 84.14\\
    & \checkmark & 73.69 & 83.58 & 73.68 & 83.58\\
    \checkmark & \checkmark &  \textbf{75.12} & \textbf{85.09} & \textbf{75.45} & \textbf{85.31}\\
    \bottomrule
    \end{tabular}
    
\end{table}

\section{EXPERIMENT}
\subsection{Setting}
\textbf{Datasets and evaluation metrics.} To evaluate P$^3$-SAM, we select three public S$^3$D benchmarks: FSSD-12 \cite{r32}, Surface Defects-4i \cite{r33}, and ESDIs-SOD \cite{r14,r34}. Following \cite{r14,r32,r35}, we evaluate segmentation performance using mean Intersection over Union (mIoU) and Foreground-Background Intersection over Union (FBIoU).

\textbf{Implementation details.} 
All experiments are conducted on a single NVIDIA RTX 4090 GPU. 
We use ResNet-50 as the image encoder, initialized with ImageNet \cite{r01} pretrained weights and SAM with a ViT-H backbone at $\times$1024 resolution.
Following standard few-shot episodic training, we employ 50 learnable queries.
The backbone is trained at a learning rate of $5 \times 10^{-5}$, with all BatchNorm layers in evaluation mode to freeze their running statistics. 
The decoder head, projection layer, and feature fusion module are trained at $1 \times 10^{-4}$. 
A single AdamW optimizer is used with weight decay $1 \times 10^{-6}$ and a cosine annealing learning rate schedule. 
The model is trained for 200 epochs with a batch size of 2.

\subsection{Comparison with the State-of-the-Art}
To validate our method's effectiveness , we compare P$^3$-SAM with six state-of-the-art approaches on FSSD-12, Surface Defects-4i, and ESDIs-SOD datasets: TGRNet \cite{r33}, CPANet \cite{r32}, SCCAN \cite{r20}, DCP \cite{r36}, VRP-SAM \cite{r18}, and MAPTNet \cite{r14}. As shown in Table~\ref{tab:comparison}, P$^3$-SAM achieves superior performance across all benchmarks, outperforming prototype-based approaches such as MAPTNet. While VRP-SAM demonstrates visual reference prompting advantages in natural scenes, it encounters challenges on S$^3$D datasets, particularly in handling low-contrast regions and preserving fine-grained textures. These comparative results confirm that our method effectively integrates both semantic and spatial guidance for stable S$^3$D segmentation. 
Our method’s superiority stems from the synergistic effect of POE and PPG strategies, which well address challenges of S$^3$D images.
Specifically, on the Surface Defects-4i dataset, which is characterized by lower contrast, denser textures, greater illumination variability, and narrower defects, P$^3$-SAM achieves notable improvements of 12.00\% in mIoU and 6.75\% in FBIoU for the 1-shot, and 12.18\% in mIoU and 6.78\% in FBIoU for the 5-shot, demonstrating that our POE and PPG strategies effectively address such challenging characteristics.
Fig.~\ref{fig:motivation} presents qualitative comparisons against MAPTNet and VRP-SAM across representative defect classes in FSSD-12. P$^3$-SAM consistently produces more accurate masks across multiple challenging scenarios: noisy backgrounds in columns 1-2, low-contrast regions in column 3, complex structures such as slag inclusions in columns 4 and 6, and fine details such as scratches in column 5. These visualizations confirm that P$^3$-SAM achieves superior segmentation performance under diverse challenging conditions of S$^3$D.

\subsection{Ablation Studies}
To validate each component's contribution, we conduct ablation experiments on FSSD-12 by incrementally adding POE and PPG strategies to the baseline and report mIoU and FBIoU metrics for both 1-shot and 5-shot in Table~\ref{tab:ablation}. 
The baseline model excludes both POE and PPG strategies, relying only on visual reference prompting to extend SAM. Introducing POE strategy improves all metrics, demonstrating that POE effectively enhances shallow, fine-grained structural and textural cues tailored to S$^3$D characteristics.
Adding PPG strategy further boosts performance, indicating that PPG's dual generation of prompts provides comprehensive guidance to SAM's decoder across varying images. Our full model achieves the best performance with improvements of 3.3\% in mIoU and 2.78\% in FBIoU for 1-shot, and 3.73\% in mIoU and 3.04\% in FBIoU for 5-shot over the baseline, confirming the effectiveness of our approach for few-shot S$^3$D segmentation.

\section{CONCLUSION}
In this paper, we propose P$^3$-SAM, a novel framework that empowers SAM with Perceptual Parallel Prompt for few-shot S$^3$D segmentation.
To address the challenges of feature encoding and effective prompting for few-shot S$^3$D, we design two strategies: POE optimizes feature representations to preserve critical low-level details, while PPG generates semantic and spatial prompts for comprehensive guidance. 
Extensive experiments demonstrate that P$^3$-SAM achieves state-of-the-art performance on three benchmarks.

\bibliographystyle{IEEEtran}   
\bibliography{IEEEabrv,myref}       

\end{document}